\documentclass{article}

\usepackage{arxiv}
\usepackage{natbib}
\setcitestyle{numbers,square}
\usepackage[utf8]{inputenc} 
\usepackage[T1]{fontenc}    
\usepackage{hyperref}       
\usepackage{url}            
\usepackage{booktabs}       
\usepackage{amsfonts}       
\usepackage{nicefrac}       
\usepackage{microtype}      
\usepackage{lipsum}
\usepackage{graphicx}
\usepackage{amsmath}
\usepackage{indentfirst}
\graphicspath{ {./images/} }

\title{SASP: Strip-Aware Spatial Perception for Fine-Grained Bird Image Classification}

\author{
 Zheng Wang\\
  Central China Normal University\\
  Wuhan, China \\
  \texttt{zhengw@mails.ccnu.edu.cn} \\}
\date{}

\begin{document}
\maketitle

\begin{abstract}
Fine-grained bird image classification (FBIC) is not only of great significance for ecological monitoring and species identification, but also holds broad research value in the fields of image recognition and fine-grained visual modeling. Compared with general image classification tasks, FBIC poses more formidable challenges: 1) the differences in species size and imaging distance result in the varying sizes of birds presented in the images; 2) complex natural habitats often introduce strong background interference; 3) and highly flexible poses such as flying, perching, or foraging result in substantial intra-class variability. These factors collectively make it difficult for traditional methods to stably extract discriminative features, thereby limiting the generalizability and interpretability of models in real-world applications. To address these challenges, this paper proposes a fine-grained bird classification framework based on strip-aware spatial perception, which aims to capture long-range spatial dependencies across entire rows or columns in bird images, thereby enhancing the model’s robustness and interpretability. The proposed method incorporates two novel modules: extensional perception aggregator (EPA) and channel semantic weaving (CSW). Specifically, EPA integrates local texture details with global structural cues by aggregating information across horizontal and vertical spatial directions. CSW further refines the semantic representations by adaptively fusing long-range and short-range information along the channel dimension. Built upon a ResNet-50 backbone, the model enables jump-wise connection of extended structural features across the spatial domain. Experimental results on the CUB-200-2011 dataset demonstrate that our framework achieves significant performance improvements while maintaining architectural efficiency.Thanks for the support provided by MindSpore Community.
\end{abstract}


\section{Introduction}

Fine-grained bird image classification (FBIC) plays an indispensable role in ecological research and biodiversity conservation. As sensitive bioindicators, birds’ distribution, movement patterns, and population dynamics are routinely used to assess ecosystem health. Consequently, leveraging computer vision to automate bird species identification has emerged as a vital enabler for intelligent ecological monitoring and digital conservation. With global bird populations in steady decline and the number of threatened species rising, there is an urgent need for accurate and efficient classification models. Furthermore, FBIC plays a significant role in a variety of downstream tasks and has attracted extensive attention in fields such as  image classification \citep{wuDeepModelingOptimization2025}, multimedia processing \citep{kerkouriModelingMOSQuality2025}, and bird conservation\citep{tobiasBirdConservationTropical2013}.

\par{Nevertheless, FBIC presents formidable challenges. On one hand, individuals of the same species may exhibit pronounced intra-class variability due to differences in season, age, or behavior, complicating consistent feature representation. On the other hand, inter-species distinctions are often extremely subtle—manifesting only in localized changes in plumage coloration, patterning, or posture—making discriminative learning difficult. Furthermore, complex natural backgrounds and occlusions further undermine the stability of feature extraction.}

\par{To address these difficulties, recent research has pursued diverse deep-learning–based strategies. One family of approaches focuses on localizing critical discriminative regions through region proposal and attention mechanisms to enhance sensitivity to fine-grained differences. Another line of work emphasizes global semantic understanding by integrating hierarchical feature fusion or Transformer architectures to capture long-range dependencies. Despite these advances, effectively balancing detailed local description with holistic spatial structure modeling remains an open challenge in fine-grained bird classification.}

\begin{figure}[h] 
	\centering 
	\includegraphics[scale=0.48]{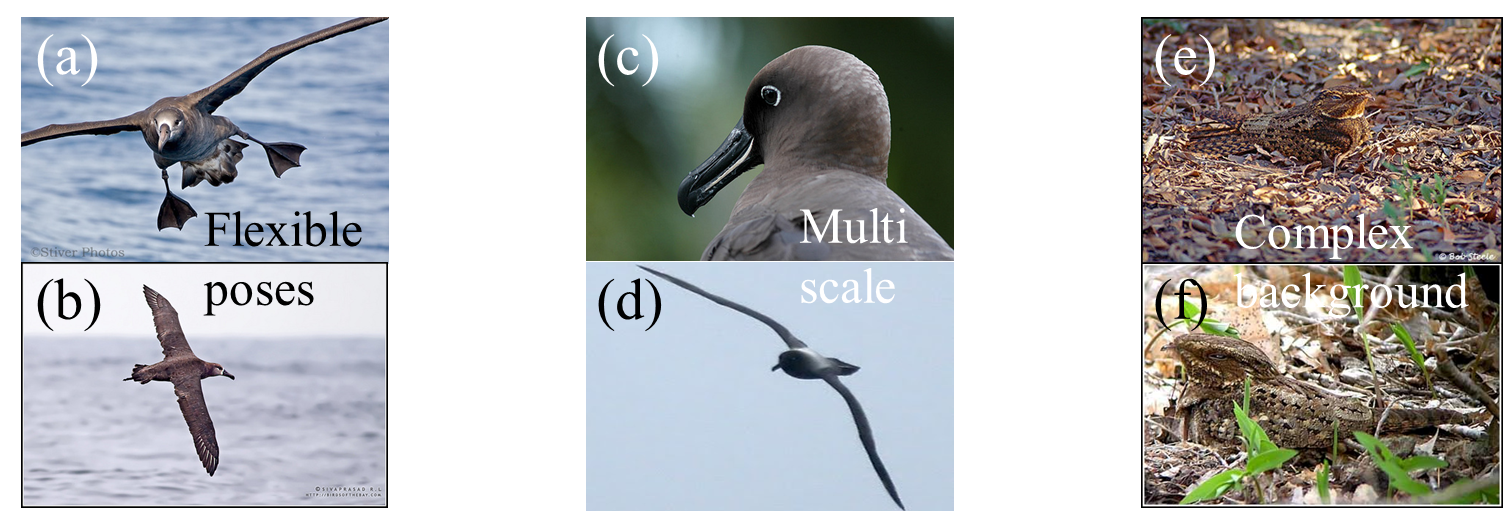}
	\caption{Challenges exist in FBIC.} 
	\label{FIG:1} 
\end{figure}

\par{Despite substantial advances in FBIC, achieving high-precision recognition remains hindered by three principal challenges (Fig. 1):}

\begin{itemize}
\item Pose Variation: Taking the Black-footed Albatross as an example, its wing-outstretched horizontal flight Fig.1(a) and vertical dive posture Fig.1(b) exhibit markedly different spatial feature distributions. Fixed receptive fields in conventional convolutional networks struggle to accommodate such directional diversity simultaneously.
\item Scale Variation: In wild settings, the proportion of a bird’s body in the image can vary dramatically. Fig.1(c) captures only the compact head of the albatross, whereas Fig.1(d) depicts its full-body wingspan at a distance. This extreme scale disparity causes fine details of small targets (e.g., plumage spots or feather patterns) to be compressed or distorted in shallow feature maps, while large targets may exceed local convolutional receptive fields, resulting in fragmented representations and undermining both intra-class consistency and multi-scale feature fusion.
\item Complex Backgrounds: Birds frequently inhabit environments with dense foliage or dappled lighting—e.g., the Chuck-will’s-widow in Fig. 1(e) and Fig.1(f) is concealed among dry branches and leaves, blending deeply into its surroundings. Such background clutter not only complicates bird localization but also misleads feature extraction modules, thereby degrading classification reliability.
\end{itemize}

\par{To address the challenges of pose variation, scale discrepancy, and background clutter in FBIC, we propose a strip-aware spatial perception for FBIC (SASP). Our model builds upon a ResNet-50 backbone truncated before its final residual stage—to produce high-dimensional feature maps. We then inject an extensional perception aggregator (EPA) at the uppermost feature level, where parallel branches of local convolution and horizontal-vertical strip pooling capture both fine-grained texture and extended structural contours. After upsampling to the original resolution, these branches are fused to enable “leap-wise” modeling of long-range dependencies across varying flight poses and object scales. A subsequent channel semantic weaving (CSW) layer performs channel-wise reweighting by first generating a compact global semantic descriptor via pooling and dimensionality reduction, then adaptively calibrating individual feature channels to amplify discriminative signals and suppress background noise. The resultant feature representation is finally subjected to global average pooling and a multi-layer perceptron head to yield predictions over 200 bird categories, delivering both high accuracy and interpretability. Extensive evaluations on CUB-200-2011 demonstrate that our method substantially outperforms baseline approaches while maintaining computational efficiency, and that the learned EPA strip responses and CSW channel weights offer valuable insights into the model’s decision process.}

\par{In summary, the major contributions of this work are listed as follows:}

\begin{itemize}
	\item We introduce a novel extensional perception aggregator that leverages parallel local convolution and horizontal-vertical strip pooling to capture long-range spatial dependencies, enabling leap-wise modeling of diverse flight poses and scale variations in bird images.
	\item We propose a channel semantic weaving layer that adaptively reweights feature channels using a compact global semantic descriptor, thereby enhancing discriminative signals while suppressing background noise for fine-grained classification.
	\item We demonstrate through experiments on CUB-200-2011 that our strip-aware spatial perception framework achieves impressive accuracy with high computational efficiency.
\end{itemize}

\par{The remainder of this paper is arranged as follows. The research problem is formally defined, and the specifics of SASP are presented in Section 2. In Section 3, the experimental results of the proposed model on CUB-200-2011 datasets are presented. The conclusions are discussed in Section 4.}

\section{Proposed SASP method}
\subsection{Overview}
\par{The overall architecture of proposed strip-aware spatial perception model is illustrated in Fig. 2. It comprises four main components: a feature extraction backbone, extensional perception aggregator (EPA), channel semantic weaving (CSW) layer, and the classification head. First, a ResNet-50 backbone—pruned before its final residual stages—is used to encode fine-grained and multiscale bird image features. We enhance this backbone by inserting dropout layers into its classification head to mitigate overfitting. Second, the EPA module operates on the high-level feature map in parallel branches: standard $3\times3$ convolutions capture local texture, while horizontal and vertical strip pooling with asymmetric convolutions aggregate long-range structural cues. These branch outputs are upsampled, fused, and added residually to the input feature map, enabling “leap-wise” modeling of pose and scale variations. Third, the CSW layer performs channel-wise reweighting by first applying global average pooling to produce a compact semantic descriptor, then passing it through a two-layer bottleneck and sigmoid activation to generate per-channel attention weights that amplify discriminative channels and suppress background noise. Finally, the reweighted feature map is reduced via global average pooling and fed into a lightweight multi-layer perceptron head, which outputs the final bird species predictions. This end-to-end pipeline balances local detail with global context and demonstrates strong robustness and interpretability across diverse FBIC benchmarks.}

\begin{figure}[h] 
	\centering 
	\includegraphics[scale=0.48]{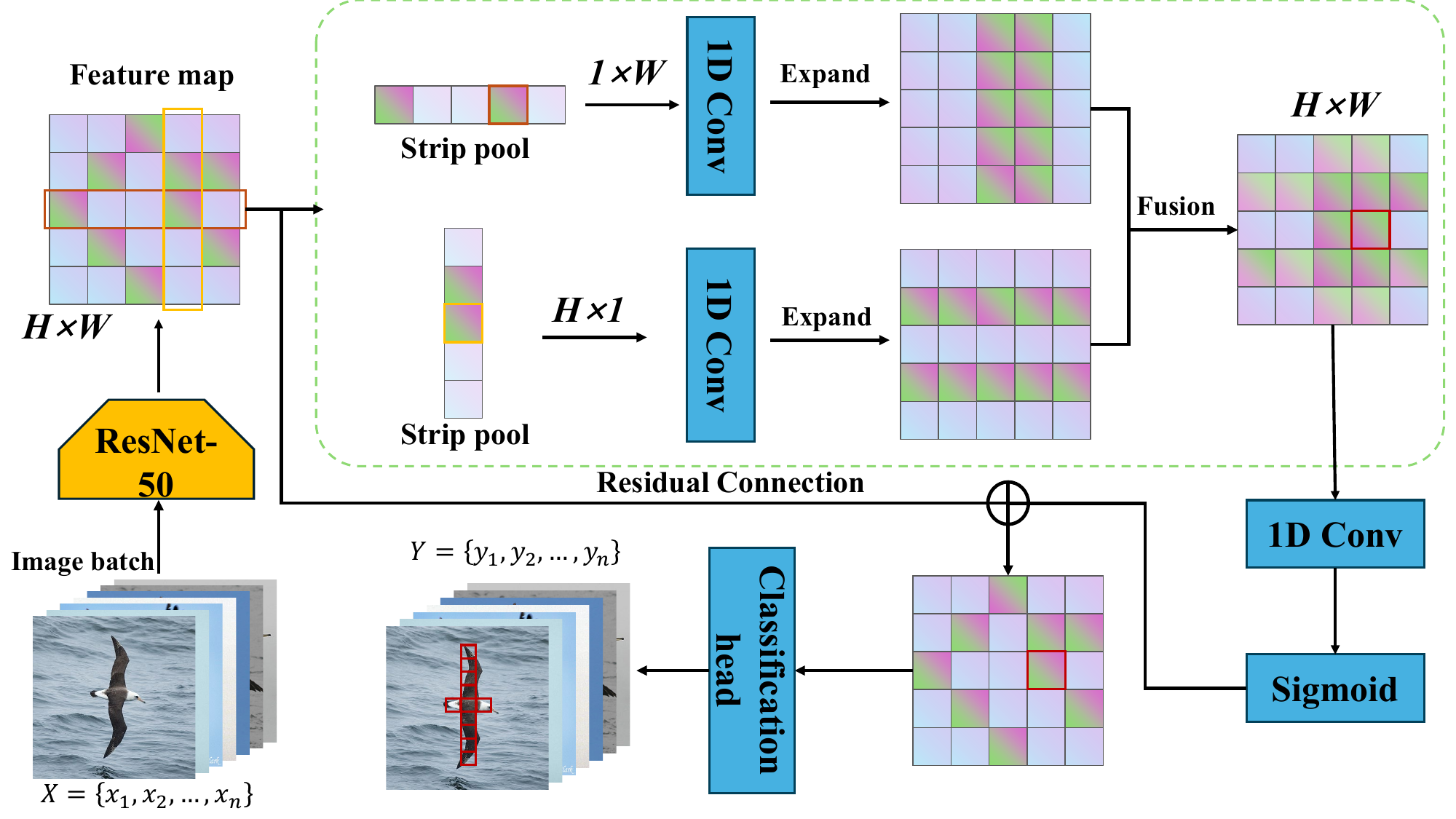}
	\caption{Overview of the proposed SASP architecture.} 
	\label{FIG:2} 
\end{figure}

\subsection{Extensional perception aggregator}
\par{Given a sequence of bird images $ X=\left\{x_1,x_2,\ldots,x_n\right\}$, we first employ ResNet-50 as the backbone encoder. Each input image $ x\in\mathbb{R}^{3\times H\times W}$ is transformed into a deep semantic feature map: 
\begin{equation}
	F=\mathcal{R}\left(x\right)\in\mathbb{R}^{C\times H^\prime\times W^\prime},
\end{equation}
where $C=2048$ and $H^\prime=W^\prime=7 $. $\mathcal{R}\left(\cdot\right)$ represents ResNet-50. This feature map preserves both local texture cues (e.g., feather spots and beak details) and mid-level structural information. However, a standard $3\times3$ convolution only captures short-range dependencies among neighboring pixels, making it inadequate for modeling long-range, strip-like structures such as wings in flight or elongated tail feathers across varying poses and scales. To address this limitation, we introduce the extensional perception aggregator, specifically designed to capture these cross-row and cross-column dependencies.
\par{The EPA is composed of five parallel branches. Specifically, the local branch applies a $3\times3$ convolution to maximally preserve high-frequency detail from neighboring pixels—thereby capturing discriminative cues such as feather patterns and spots—formulated as:}
\begin{equation}
   F_{loc}=Conv\left(F\right),
\end{equation}
where $Conv(\bullet)$ represents $3\times3$ convolution.}
\par{Next, we apply horizontal and vertical strip pooling to the feature map. First, we compress $F $ along the row dimension to obtain:}
\begin{equation}
U_h=Pool_{H^{\prime}\times1}(F)\in\mathbb{R}^{C\times H^{\prime}\times1},
\end{equation}
where ${Pool}_{H^\prime\times1}$ denotes adaptive average pooling across each row. A $3\times3$ convolution is then performed on $ U_h$, and the result is upsampled back to $ \left(H^\prime\times W^\prime\right)$ by bilinear interpolation:

\begin{equation}
F_h=Up(Conv(U_h))\in\mathbb{R}^{C\times H^{\prime}\times W^{\prime}}.
\end{equation}
\par{The horizontal strip pooling aggregates semantics across entire rows, emphasizing the contiguous, lateral extension of structures such as outstretched wings or tail feathers. Analogously, we perform vertical strip pooling by compressing $F$ along the column dimension:
\begin{equation}
U_v={Pool}_{1\times W^\prime}\left(F\right)\in\mathbb{R}^{C\times1\times W^\prime},
\end{equation}
\begin{equation}
F_v=Up\left(Conv\left(U_v\right)\right)\in\mathbb{R}^{C\times H^\prime\times W^\prime}.
\end{equation}
}
\par{This branch captures the longitudinal spine of the bird from head to tail. To further enrich directional feature dynamics, we first reduce the channel dimensionality of $F$ via a $1\times1$ convolution, producing $ {F^\prime\in\mathbb{R}}^{C^\prime\times H^\prime\times W^\prime}\ (C^\prime=C/4).$ We then refine horizontal and vertical strip features using asymmetric kernels:
\begin{equation}
    F_{sh}=Up\left({Conv}_{1\times3}\left({Pool}_{1\times W^\prime}\left(F^\prime\right)\right)\right),
\end{equation}
\begin{equation}
	F_{sv}=Up\left({Conv}_{3\times1}\left({Pool}_{H^\prime\times1}\left(F^\prime\right)\right)\right).
\end{equation}}
\par{So that it flexibly models continuous texture flows along strip directions. Finally, to achieve robust multi-scale compatibility, we fuse all five branch outputs. We first compute two intermediate fused features:}
\begin{equation}
    F^{\left(1\right)}=\sigma\left(F_{loc}+F_h+F_v\right),
\end{equation}
\begin{equation}
	F^{\left(2\right)}=\sigma\left(F_{sh}+F_{sv}\right).
\end{equation}
where $\sigma\left(\cdot\right)$ denotes ReLU activation. These are then concatenated along the channel axis and projected back to the original channel dimension by a $1\times1$ convolution:
\begin{equation}
    F_{EPA}={Conv}_{1\times1}\left(\left[F^{\left(1\right)},F^{\left(2\right)}\right]\right)\in\mathbb{R}^{C\times H^\prime\times W^\prime}.
\end{equation}
\par{A residual connection adds the original feature map $F$ to $F_{EPA}$, followed by a final activation:}
\begin{equation}
	F_{out}=\sigma\left(F+F_{EPA}\right).
\end{equation}
\par{Through this five-branch architecture, the EPA module preserves local detail while effectively capturing and fusing long-range, cross-row and cross-column dependencies. It substantially expands the network’s receptive field and endows the model with “leap-wise” perception of strip-like structures—such as fan-shaped wings and elongated tail feathers—across diverse poses and scales, thereby providing high-quality features for subsequent channel semantic weaving and final classification.}
\subsection{Channel Semantic Weaving}
\par{After receiving the enhanced feature map $F_{out}\ \in\mathbb{R}^{C\times H^\prime\times W^\prime}$ from the EPA module, the CSW layer performs channel-wise recalibration to emphasize informative feature channels and suppress noise. First, a compact global descriptor $s\in\mathbb{R}^C$ is computed via spatial average pooling:}

\begin{equation}
s=GAP\left(F_{out}\right)=\frac{1}{H^\prime W^\prime}\sum_{i=1}^{H^\prime}\sum_{j=1}^{W^\prime}{F_{out}\left[:,i,j\right]}.
\end{equation}
    
\par{Then the vector is passed through a two-stage bottleneck: }
\begin{equation}
\begin{cases}
	z=W_1s+b_1,\\
	\widetilde{z}=ReLU\left(z\right),\\
	w=W_2\widetilde{z}+b_2, & 
\end{cases}
\end{equation}
where $W_1\ \in\mathbb{R}^{C/r\times C}$ reduces the channel dimension by factor $r$, and $W_1\ \in\mathbb{R}^{C\times C/r}$ restores it. A sigmoid activation $\alpha=\sigma\left(w\right)\in\left(0,1\right)^C$ produces per-channel attention weights, which are then broadcast across spatial positions and applied to reweight the feature map:
\begin{equation}
	F_{CSW}=\ \alpha\ \odot\ F_{out},
\end{equation}
\par{This mechanism dynamically amplifies the most informative channels such as those corresponding to a bird’s beak or crest while suppressing noise channels, thereby enabling adaptive, lightweight channel recalibration that complements the long- and short-range spatial cues captured by the EPA module.}

\subsection{Training Loss}
The output of the CSW layer $F_{CSW}\ \in\mathbb{R}^{C\times H^\prime\times W^\prime}$ is then reduced to a compact feature vector via global average pooling:
\begin{equation}
v=GAP\left(F_{CSW}\right)\in\mathbb{R}^C,
\end{equation}
which is fed into a lightweight multi-layer perceptron classifier. Specifically, we apply two successive ReLU-activated linear transformations followed by a final affine projection to produce the class logits $z\in\mathbb{R}^K$:
\begin{equation}
	z=W_3ReLU\left(W_2ReLU\left(W_1v+b_1\right)+b_1\right)+b_3.
\end{equation}
\par{During training, the network parameters are optimized by minimizing the standard cross-entropy loss between the predicted logits and the one-hot ground-truth labels $y$,}
\begin{equation}
	\mathcal{L}=-\sum_{k=1}^{K}{y_klog\left(\phi\left(z_k\right)\right)},
\end{equation}
where $\phi\left(\cdot\right)$ denotes $softmax(\cdot)$. This end-to-end learning objective ensures that spatial cues captured by EPA and channel weights learned by CSW are jointly tuned to maximize fine-grained bird classification performance.
\section{Experimental results and discussion}
\par{In this section, we provide a detailed introduction to the  dataset used in our experiments, followed by the experimental setup and a comprehensive presentation of the results. The evaluation includes comparisons with other methods, along with visualizations that demonstrate the effectiveness and predictive performance of our model.}
\subsection{Dataset}
\par{The CUB-200-2011 \citep{wahCaltechucsdBirds2002011Dataset2011a} dataset comprises 200 bird species with a total of 11,788 images, of which 5,994 are allocated for training and 5,794 for testing. Sample images from this dataset are illustrated in Fig. 3. Because the birds depicted are generally small in size and inter-species distinctions often hinge on extremely subtle variations in plumage patterns, spots, or posture, extracting and discriminating these fine-grained features presents a particularly formidable challenge.} 

\begin{figure}[h] 
	\centering 
	\includegraphics[scale=0.48]{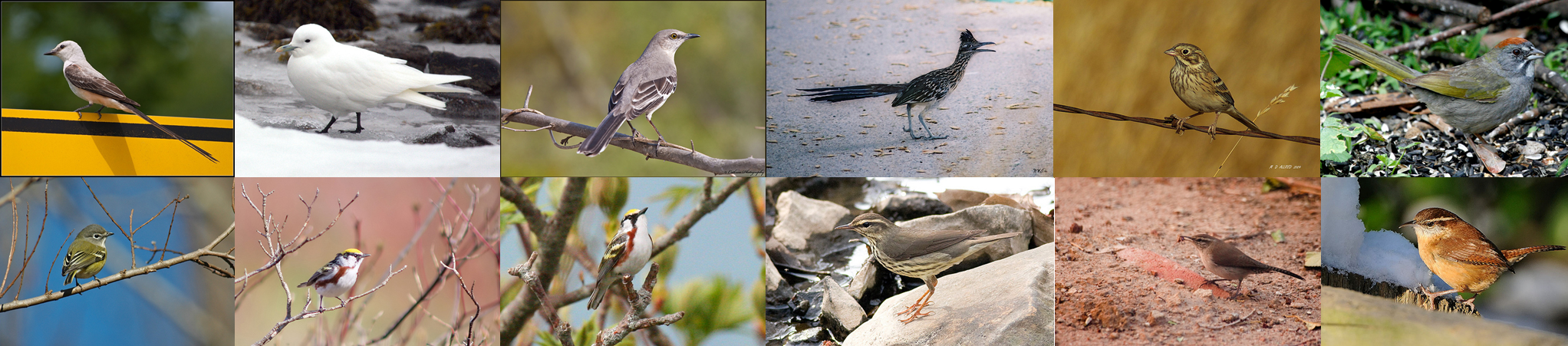}
	\caption{Samples in CUB-200-2011 dataset.} 
	\label{FIG:3} 
\end{figure}

\subsection{Implementation Details}
\par{In our experimental protocol, facial regions are first automatically detected and cropped to ensure fair comparison, after which all crops are uniformly resized to a resolution of  $224\times224$. The model is trained for 70 epochs with a batch size of 256. We employ a Momentum optimizer initialized with a learning rate of 0.1, which is subsequently reduced to 0.01 using a polynomial decay schedule with a power of 0.5. Experiments are conducted on a computing platform equipped with an NVIDIA T4 GPU. All model development and training procedures are implemented in the MindSpore framework. The detailed experimental parameter Settings are shown in Table 1.}
\subsection{Experiment Results and Analysis}	
\par{As shown in Table 2, our strip-aware spatial perception (SASP) model achieves a classification accuracy of 72.58\% on the CUB-200-2011 dataset based on MindSpore framework, demonstrating a substantial performance gain. In contrast, GoogLeNet attains only 68.19\%, significantly underperforming our SASP framework. Although GoogLeNet’s Inception modules extract multi-scale local features in parallel at each layer—thereby capturing certain fine-grained texture variations—its receptive field and channel interaction mechanisms remain constrained by fixed convolutional kernel configurations, limiting its ability to model long-range cross-row and cross-column dependencies and undermining robustness to complex poses and large-scale variations. Furthermore, GoogLeNet lacks an explicit channel recalibration strategy, treating all channel outputs equally and failing to suppress background clutter and subtle feather-pattern noise. By comparison, our EPA module employs strip pooling to “leap-wise” extend the receptive field and aggregate both short- and long-range spatial cues, while the CSW layer adaptively amplifies critical channels and attenuates redundant signals to precisely capture ultra-fine-grained differences. These design innovations enable SASP to substantially overcome the challenges of pose variation, scale disparity, and complex backgrounds, thereby delivering superior classification performance.}

\begin{table}[h]
	\centering
	\caption{Experimental parameter settings.}
	\label{tab:model_comparison}
	\footnotesize
	\begin{tabular}{@{} c c c c c c c @{}}
		\toprule
		\textbf{Dataset} & \textbf{Batch size} &\textbf{Learning rate}& \textbf{Picture resolution}& \textbf{In channels}& \textbf{Epochs}& \textbf{Weight decay} \\ 
		\midrule
		CUB-200-2011&256&0.1&$224\times224$&2048&70&1e-4\\
		\bottomrule
	\end{tabular}
	
\end{table}

\begin{table}[!h]
	\centering
	\caption{Experiment results based on MindSpore framework. Comparison with the state-of-the-art results on the CUB-200-2011 dataset. The best results are in \textbf{BOLD}, and the second-best results are \underline{underlined}.}
	\label{tab:model_comparison}
	\footnotesize
	\begin{tabular}{@{} c c c @{}}
		\toprule
		\textbf{Model} & \textbf{Proc.}  & \textbf{Acc. (\%)} \\ 
		\midrule
		
		Berg et al. \citep{t.bergPOOFPartBasedOnevs2013} & ICCV  & 56.78 \\ 
		Goering et al. \citep{c.goeringNonparametricPartTransfer2014}&CVPR & 57.84 \\ 
		Chai et al. \citep{y.chaiSymbioticSegmentationPart01a}& ICCV & 59.40 \\ 		 
		Simon et al. \citep{simonPartDetectorDiscovery2014}& ACCV& 62.53\\
		Donahue et al. \citep{donahueDecafDeepConvolutional2014}&ICML & 64.96\\
		Xiao et al. \citep{xiaoApplicationTwolevelAttention2015}&CVPR&\underline{69.70}\\
		C. Szegedy et al. \citep{c.szegedyGoingDeeperConvolutions2015}& CVPR &68.19\\
		SASP (OURS) & - &\textbf{72.58} \\ 
		\bottomrule
	\end{tabular}
	\vspace{-3ex}
\end{table}

\subsection{Visualization}	

\begin{figure}[h] 
	\centering 
	\includegraphics[scale=0.9]{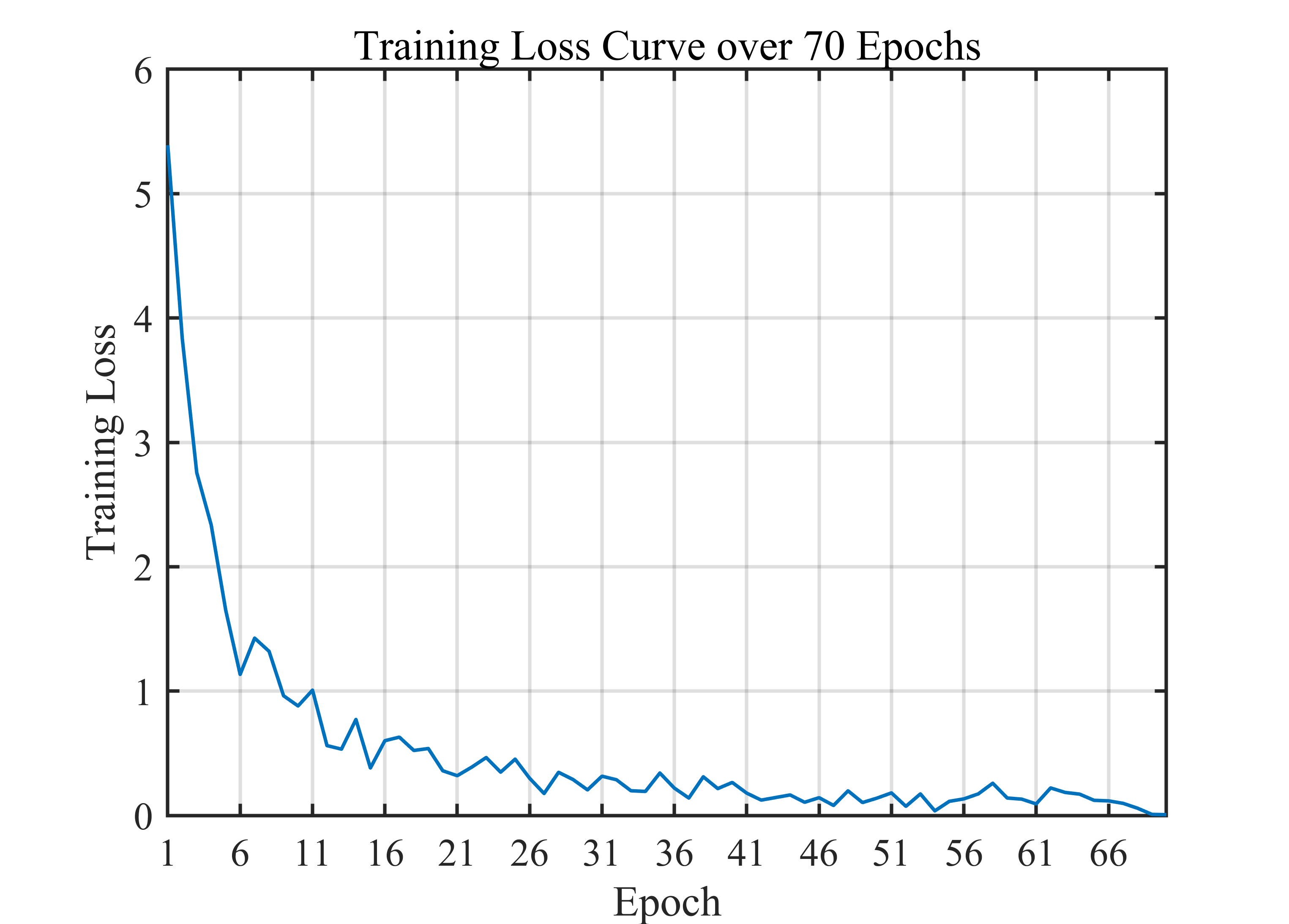}
	\caption{Training loss over 70 epochs.} 
	\label{FIG:4} 
\end{figure}

\par{As shown in Figure 4, our loss curve exhibits a rapid decline in the early stages, indicating that the EPA module effectively captures both short- and long-range spatial dependencies; it then stabilizes during mid-training, reflecting how the CSW layer incrementally refines channel-wise recalibration to enhance discriminative power. Finally, with the aid of a polynomial learning rate decay, the model undergoes fine-tuning, maintaining a steady downward trend and ultimately achieving efficient convergence and robust classification of subtle texture variations.
}

\section{Conclusion}
\par{In this work, we introduce a novel fine-grained bird classification framework underpinned by strip-aware spatial perception, which excels at modeling long-range structural cues and discriminative texture details. Our EPA module “leap-wise” captures both local and global strip-like patterns—such as outstretched wings and elongated tail feathers—by fusing standard convolutions with horizontal and vertical strip pooling. Complementarily, the CSW layer dynamically recalibrates channel responses, amplifying key semantic features while suppressing background noise. Comprehensive experiments on CUB-200-2011 demonstrate that our approach significantly achieves significant performance improvements. These results confirm that synergistically combining long-range spatial dependencies with adaptive channel attention markedly enhances robustness and precision in the face of pose variation, scale disparity, and complex backgrounds.}

\section*{Acknowledgments}
\par{Thanks for the support provided by MindSpore Community. All experiments proposed in this paper are imple mented based on the mindspore framework.}

\bibliographystyle{unsrt}  

\bibliography{SASP.bib}  

\begin{thebibliography}{10}

\bibitem{wuDeepModelingOptimization2025}
Yihang Wu, Muhammad Owais, Reem Kateb, and Ahmad Chaddad.
\newblock Deep {{Modeling}} and {{Optimization}} of {{Medical Image
  Classification}}.
\newblock In {\em 2025 {{IEEE}} 22nd {{International Symposium}} on
  {{Biomedical Imaging}} ({{ISBI}})}, pages 1--4. IEEE.

\bibitem{kerkouriModelingMOSQuality2025}
Mohamed~Amine Kerkouri, Marouane Tliba, Aladine Chetouani, Nour Aburaed, and
  Alessandro Bruno.
\newblock Modeling {{Beyond MOS}}: {{Quality Assessment Models Must Integrate
  Context}}, {{Reasoning}}, and {{Multimodality}}.

\bibitem{tobiasBirdConservationTropical2013}
Joseph~A. Tobias, Çağan~H. Şekercioğlu, and F.~Hernan Vargas.
\newblock Bird conservation in tropical ecosystems: Challenges and
  opportunities.
\newblock pages 258--276.

\bibitem{wahCaltechucsdBirds2002011Dataset2011a}
Catherine Wah, Steve Branson, Peter Welinder, Pietro Perona, and Serge
  Belongie.
\newblock The caltech-ucsd birds-200-2011 dataset.

\bibitem{t.bergPOOFPartBasedOnevs2013}
{T. Berg} and {P. N. Belhumeur}.
\newblock {{POOF}}: {{Part-Based One-vs}}.-{{One Features}} for {{Fine-Grained
  Categorization}}, {{Face Verification}}, and {{Attribute Estimation}}.
\newblock In {\em 2013 {{IEEE Conference}} on {{Computer Vision}} and {{Pattern
  Recognition}}}, pages 955--962.

\bibitem{c.goeringNonparametricPartTransfer2014}
{C. Göering}, {E. Rodner}, {A. Freytag}, and {J. Denzler}.
\newblock Nonparametric {{Part Transfer}} for {{Fine-Grained Recognition}}.
\newblock In {\em 2014 {{IEEE Conference}} on {{Computer Vision}} and {{Pattern
  Recognition}}}, pages 2489--2496.

\bibitem{y.chaiSymbioticSegmentationPart01a}
{Y. Chai}, {V. Lempitsky}, and {A. Zisserman}.
\newblock Symbiotic {{Segmentation}} and {{Part Localization}} for
  {{Fine-Grained Categorization}}.
\newblock In {\em 2013 {{IEEE International Conference}} on {{Computer
  Vision}}}, pages 321--328.

\bibitem{simonPartDetectorDiscovery2014}
Marcel Simon, Erik Rodner, and Joachim Denzler.
\newblock Part detector discovery in deep convolutional neural networks.
\newblock In {\em Asian Conference on Computer Vision}, pages 162--177.
  Springer.

\bibitem{donahueDecafDeepConvolutional2014}
Jeff Donahue, Yangqing Jia, Oriol Vinyals, Judy Hoffman, Ning Zhang, Eric
  Tzeng, and Trevor Darrell.
\newblock Decaf: {{A}} deep convolutional activation feature for generic visual
  recognition.
\newblock In {\em International Conference on Machine Learning}, pages
  647--655. PMLR.

\bibitem{xiaoApplicationTwolevelAttention2015}
Tianjun Xiao, Yichong Xu, Kuiyuan Yang, Jiaxing Zhang, Yuxin Peng, and Zheng
  Zhang.
\newblock The application of two-level attention models in deep convolutional
  neural network for fine-grained image classification.
\newblock In {\em Proceedings of the {{IEEE}} Conference on Computer Vision and
  Pattern Recognition}, pages 842--850.

\bibitem{c.szegedyGoingDeeperConvolutions2015}
{C. Szegedy}, {Wei Liu}, {Yangqing Jia}, {P. Sermanet}, {S. Reed}, {D.
  Anguelov}, {D. Erhan}, {V. Vanhoucke}, and {A. Rabinovich}.
\newblock Going deeper with convolutions.
\newblock In {\em 2015 {{IEEE Conference}} on {{Computer Vision}} and {{Pattern
  Recognition}} ({{CVPR}})}, pages 1--9.

\end{thebibliography}

\end{document}